\documentclass[letterpaper, 10 pt, conference]{ieeeconf}

\IEEEoverridecommandlockouts               

\usepackage[style=ieee]{biblatex}
\usepackage{amsmath}
\usepackage{graphicx}
\usepackage{subcaption}
\usepackage{url}
\usepackage{csquotes}
\usepackage[american]{babel}
\usepackage{hyperref}
\usepackage{textcomp}
\usepackage{tikz}
\newcommand\copyrighttext{%
  \footnotesize \textcopyright 2020 IEEE. Personal use of this material is permitted.
  Permission from IEEE must be obtained for all other uses, in any current or future
  media, including reprinting/republishing this material for advertising or promotional
  purposes, creating new collective works, for resale or redistribution to servers or
  lists, or reuse of any copyrighted component of this work in other works.
  DOI: \href{https://doi.org/10.1109/ICRA40945.2020.9196854}{10.1109/ICRA40945.2020.9196854}}
\newcommand\copyrightnotice{%
\begin{tikzpicture}[remember picture,overlay]
\node[anchor=south,yshift=10pt] at (current page.south) {\fbox{\parbox{\dimexpr\textwidth-\fboxsep-\fboxrule\relax}{\copyrighttext}}};
\end{tikzpicture}%
}

\title{PuzzleFlex: kinematic motion of chains with loose joints}
\author{Samuel Lensgraf, Karim Itani, Yinan Zhang, Zezhou Sun, Yijia Wu,\\%
Alberto Quattrini Li, Bo Zhu, Emily Whiting, Weifu Wang, Devin Balkcom
\thanks{This project was partially supported by NSF Grant 1813043, as well as by the  Dubai Future Foundation.}
}

\newcommand{\bo}{\mathbf o}
\newcommand{\bq}{\mathbf q}
\newcommand{\bp}{\mathbf p}
\newcommand{\bd}{\mathbf d}
\newcommand{\bn}{\mathbf n}
\newcommand{\bc}{\mathbf c}

\begin{document}
\maketitle
\copyrightnotice

\begin{abstract}
  This paper presents a method of computing free motions of a planar assembly of rigid bodies connected by loose joints. Joints are modeled using local distance constraints, which are then linearized with respect to configuration space velocities, yielding a linear programming formulation that allows analysis of systems with thousands of rigid bodies. Potential applications include analysis of collections of modular robots, structural stability perturbation analysis, tolerance analysis for mechanical systems, and formation control of mobile robots.
\end{abstract}

\section{Introduction}

Like a human skeleton, structures assembled by or out of robots may be composed of rigid bodies loosely connected at the joints. A many-jointed robot arm flexes like the backbone of a snake; a wooden jigsaw puzzle may flex slightly as one edge is pulled, particularly before assembly is complete. Joints may be real or virtual: enforced by the physics of collision, or by robot control laws that prevent the breaking of formation.

\begin{figure}
  \begin{subfigure}[t]{0.23\textwidth}
  	\centering
    \includegraphics[width=1.4in]{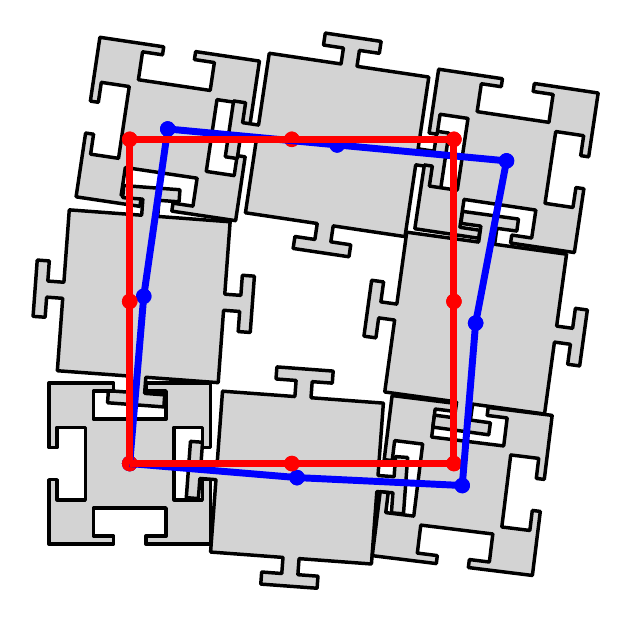}
  \end{subfigure}
  ~
  \begin{subfigure}[t]{0.23\textwidth}
    \centering
    \includegraphics[width=1.4in]{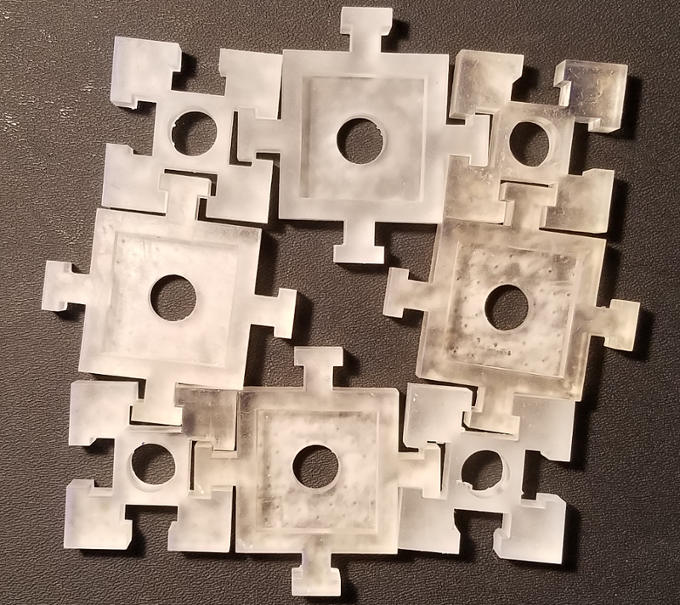}
  \end{subfigure}
  \caption{A system of eight rigid blocks (left), with the upper-right block flexing to the right.}
  \label{fig:intro_figs}
\end{figure}

This paper studies a model of the kinematics of collections of rigid bodies that are flexible in the aggregate. It presents a simple, fast, linearized method to quickly estimate potential motions of the system that maximize deviation from the initial configuration in a considered direction. First, a set of linear constraints is derived that approximates the shape of the local configuration space; then linear programming is used in various ways to optimize or analyze potential motions of the system. Figure~\ref{fig:intro_figs} shows an example of a planar puzzle flexing in such a way that the upper right block moves maximally in the positive $x$ direction. Because of the linearization, there is some violation of the constraints; the paper presents time-stepping and other methods to verify the estimate while respecting constraints.


Flexibility analysis may enable wise design decisions about robot systems or about structures that robots build. Flexibility may be good, allowing compliance with external forces, or bad, reducing the sturdiness and predictability of the system. What joint tolerances enable assembly, while providing either enough flexibility for Lego-like bricks or modular robots to comply to an external structure, or enough rigidity for the robots to resist external loads? What arrangements of bodies provide the desired level of flexibility? How much motion, and in which direction, can a system of flocking robots achieve while maintaining constraints such as mutual visibility?

\begin{figure}
  \centering
  \begin{subfigure}[t]{0.22\textwidth}
  	\centering
  	\includegraphics[height=1.7in]{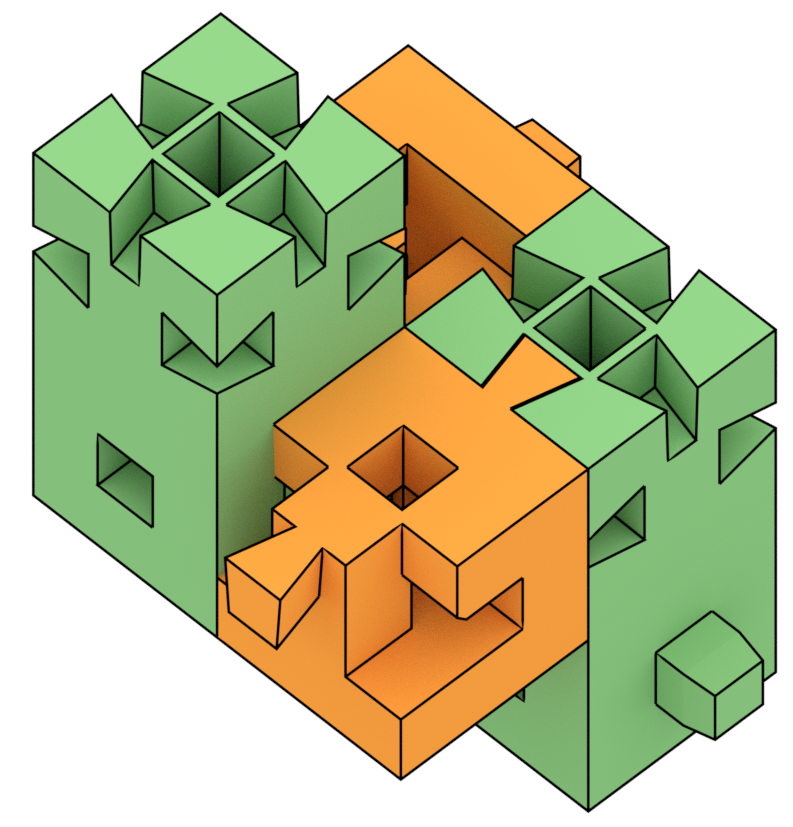}
  \end{subfigure}
  ~
  \begin{subfigure}[t]{0.22\textwidth}
  	\centering
  	\includegraphics[height=1.7in]{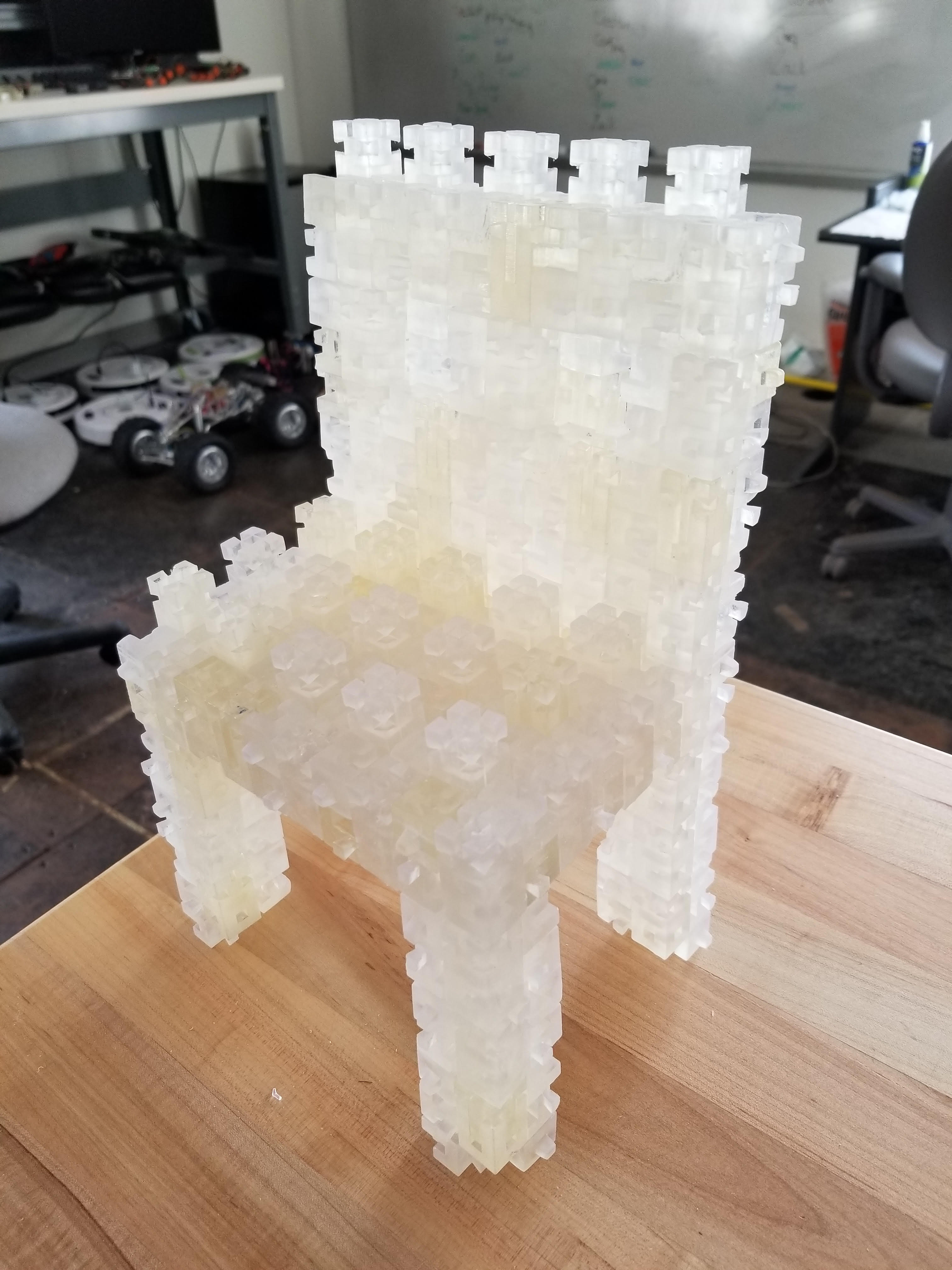}
  \end{subfigure}
\caption{Interlocking puzzle blocks, from~\cite{Zhang2018-interlocking}.}
\label{fig:chair}
\end{figure}

The work is motivated by methods for building modular interlocking structures such as those presented by Zhang {\em et al.}~\cite{Zhang2018-interlocking,Zhang2016a} and by Werfel {\em et al.}~\cite{Werfel2006-mobile-robot-construction}. Figure~\ref{fig:chair} shows a system of particular interest to the authors: a chair created from Lego-like blocks held together by a puzzle-like arrangement, rather than by friction or glue. We imagine constructing such structures automatically with robots, or from systems of modular robots. The chair flexes slightly, but remains as a single component as long as the final block assembled is held in place.
Fast analysis of flexibility will allow specific design decisions: reinforcing the chair by adding other blocks, or increasing tolerance at joints to allow easier manufacturing and assembly, while maintaining acceptable rigidity.
For simplicity, this paper focuses on planar systems.

\begin{figure}
  \centering
  \includegraphics[height=2.5in]{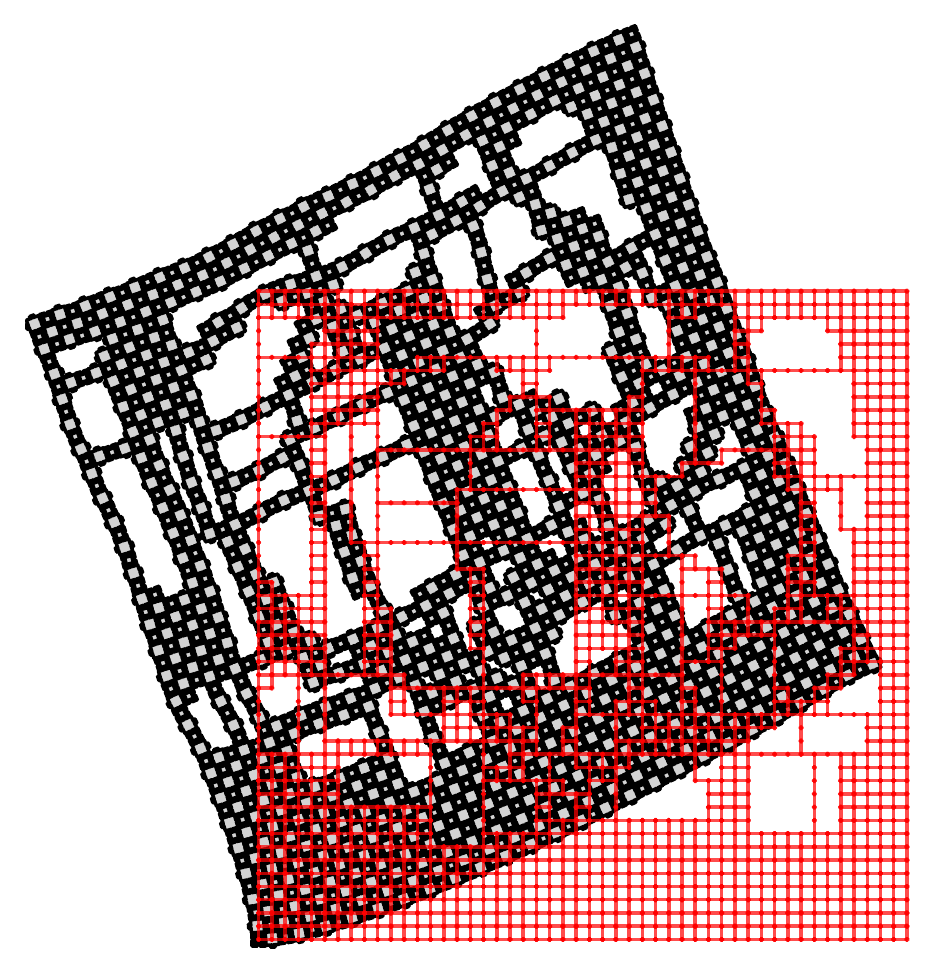}
    \caption{Structure composed of 1703 rigid bodies flexing upwards. Red polygons denote the initial configuration.}
  \label{fig:very_large_chain}
\end{figure}

We developed a Julia library to build the linear constraint matrix based on geometric descriptions of part geometry\footnote{Software available from \url{rlab.cs.dartmouth.edu}}, and used Gurobi~\cite{gurobi} to solve linear programs. Though we compute and present distance function gradients in the paper for reference, we used automatic differentiation in the implementation for simplicity~\cite{Revels2016-automatic-differentiation-in-julia}. Figure~\ref{fig:very_large_chain} shows a structure with 1703 blocks, with $y$ displacement of the upper right block maximized using this implementation.

While the present work makes use of sparse linear programming to solve for maximal motions in a direction in the configuration space of the chain, the convex polytope that approximates the shape of the local configuration space is interesting in itself. We note that global optimization techniques over the configuration space without first estimating the local constraint region, are in some ways too strong -- they may find solutions that are not connected to the initial configuration space, and are thus not reachable. Ultimately, we would like to move beyond optimizaton and explore ways of measuring properties of this polytope, and use these properties to automatically design structures, in much the same way as the manipulablity elipsoid constructed from the Jacobian has long been used in robot arm design~\cite{Chiacchio1991-manipulability,Park1998-manipulability,Kim1998-manipulability,Bicchi2000-manipulability}.

\section{Related work}
A slightly extended version of this paper is available as a technical report~\cite{lensgraf2019puzzleflex}.
The work closest to the present in spirit is on compaction of planar polygons~\cite{Li1993}; the present work differs in assumptions in allowing rotation of the polygons, motivating a linear-algebraic approach.

Flexibility analysis and simulation of continuous materials such as cloth~\cite{Baraff00}, string~\cite{Pai2002,Berenson2013-deformable}, and flexible volumes~\cite{James2003} has a rich history. Models may include finite elements (e.g.~\cite{Choi02}), or may be inherently spatially continuous (e.g. Cosserat models~\cite{Rubin2000}). The present work differs in that the component modules are rigid and of irregular shape, requiring explicit consideration of the possible configurations of the chain.

The motions in the present paper allow points and edges to approach, while balancing the rates so as to optimize net motion in some direction. The distance constraints are similar to those used in recent motion planning work~\cite{Hauser2018}, as well as in Linear Complementarity Problem (LCP) formulations of dynamics~\cite{STsiam97,TTPrs01, Balkcom2002b} and in study of The Carpenter's Rule Problem~\cite{Connelly2003-carpenters}.

Linearizing motion around an initial configuration allows for the study of systems of blocks with many thousand degrees of freedom; our approach draws inspiration from early linear grasp analysis techniques~\cite{Reuleaux1876,Mishra1987-grasp-existence}. In contrast to manipulability and grasping problems, the blocks which we consider are only loosely connected. Caging grasps~\cite{RodriguezMF12,makita2008,vahedi2008caging,erickson2003capturing,rimon1996caging,allen2015two,Makita2017-caging-survey} study how robot hands may loosely capture an object; the present paper studies motion of structures in which either pairs of blocks or combinations of many blocks may cage each other. Direct construction of configuration spaces of pairs of blocks has a long history; Sacks {\em et al.}~\cite{SacksBM17} provides a recent approach, and gives a much higher-fidelity representation of the free motions of small numbers of blocks than our edge/point distance function model. Eckstein {\em et al.}~\cite{Eckenstein2017-acceptance-area-connectors} analyze how forgiving a connector design is using an approximation of the configuration space of the joint.

Tolerance analysis of mechanical assemblies is utilized in mechanical engineering to determine how frequently small manufacturing errors in the component parts of an assembly will result in unacceptable deviations in the final assembly~\cite{Chase1991-survey-of-tolerance-analysis}. The Direct Linearization Method~\cite{Chase1996-geometric-tolerance-analysis} linearizes the homogeneous transformation matrices describing the kinematics of an assembly, and applies statistical techniques to determine what percentage of assemblies are able to be assembled. The Jacobian method and other related methods for tolerance analysis~\cite{Laperriere-jacobian-tolerance,Zuo-jacobian-torsor-tolerance} models a mechanical assembly using a set of virtual joints between each element of the kinematic chain representing the assembly.

Once the local configuration space of the chain has been modeled, we solve linear programs to analyze motion; effectively, this is a line search method for numerical optimization~\cite{powell-direct-search-optimization}. Unlike most algorithms for numerical optimization, our method finds a feasible \textit{path} through configuration space rather than a single point, since the constraint polyhedron is convex. In the  simplex method~\cite{Nelder-Mead-simplex-optimization}, an $n$-dimensional simplex is constructed that satisfies the constraints and is used as a domain for the next guess in each step, but it is not guaranteed that the path through parameter space taken from the initial guess to the final solution is entirely within feasible space.


Although swarms are not the focus of this work, we briefly explore an example of how the technical approach can be used to find motions for swarms of planar polygon robots. Techniques for robot swarm control typically must handle thousands of simple robots collectively performing some tasks, e.g., object transport~\cite{alonso2017multi}, shape generation~\cite{hsieh2008decentralized}, self-assembly~\cite{o2014self,rubenstein2014programmable}, and network connectivity~\cite{esposito2006maintaining}; perhaps the closest work in spirit to the present is~\cite{ShahrokhiMB17}, which controls swarms of robots by allowing robots to bounce off of frictionless walls.

\section{Linearized distance functions}

Let the configuration of the chain be given by $\bq \in Q$. Define two types of points of interest: vertices of the polygons describing each body in the chain $\bo(\bq)$ and collision points $\bp(\bq)$. Define a vector of signed distance functions that represents the distance of each collision point from its neighboring edges: $\bd(\bo, \bp)$. Components of the vector $\bd$ will be notated by $d_{i, j}$, where $i$ is the index of the edge and $j$ is the index of the point. To enforce that there are no collisions, $\bd(\bq) \ge 0$. (We abuse the notation to write the distance function in terms of the configuration  as both $\bo$, $\bp$ are functions of $\bq$.)

To analyze legal motion and legal nearby configurations of the chain, we may consider the configuration to be a function of time: $\bq(t)$. Let $\dot \bq \in TQ$ be a configuration-space direction indicating possible motion of the system. The instantaneous rate of change of the distance function is
\begin{equation}
  \label{eq:jacobian_d}
  \dot \bd(\bq, \dot \bq) = J_d(\bq) \dot \bq,
\end{equation}
where $J_d$ is the Jacobian of the distance function. For a small enough time step $\Delta t$, an Euler step approximates the change in distances:
\begin{equation}
  \label{eq:timestep}
  \Delta \bd(t) \approx \Delta t \dot \bd(\bq, \dot \bq).
\end{equation}

Let $\bd_0 = \bd(\bo_0, \bp_0)$ be the distances computed at the initial configuration. We would like to choose motions such that the change in distances from each collision point to each edge does not cause collision: $\Delta \bd(t) \le \bd_0$. Combining with Equations~\ref{eq:jacobian_d} and~\ref{eq:timestep},
\begin{equation}
  \label{eq:polyhedron}
   J_d(\bq) \dot \bq + \bd_0 \ge 0.
\end{equation}
The scalar $\Delta t$ has been dropped, since we may equivalently linearly scale $\dot \bq$ and scale time units such that $\Delta t = 1$. With this time scaling, the change of $\bq$ over a time step is approximated by $\Delta \bq = \dot \bq$. Thus, Equation~\ref{eq:polyhedron} bounds the change in configuration to a polyhedron.

\section{A simple example}

\begin{figure}
  \centering
  \includegraphics[width=2.4in]{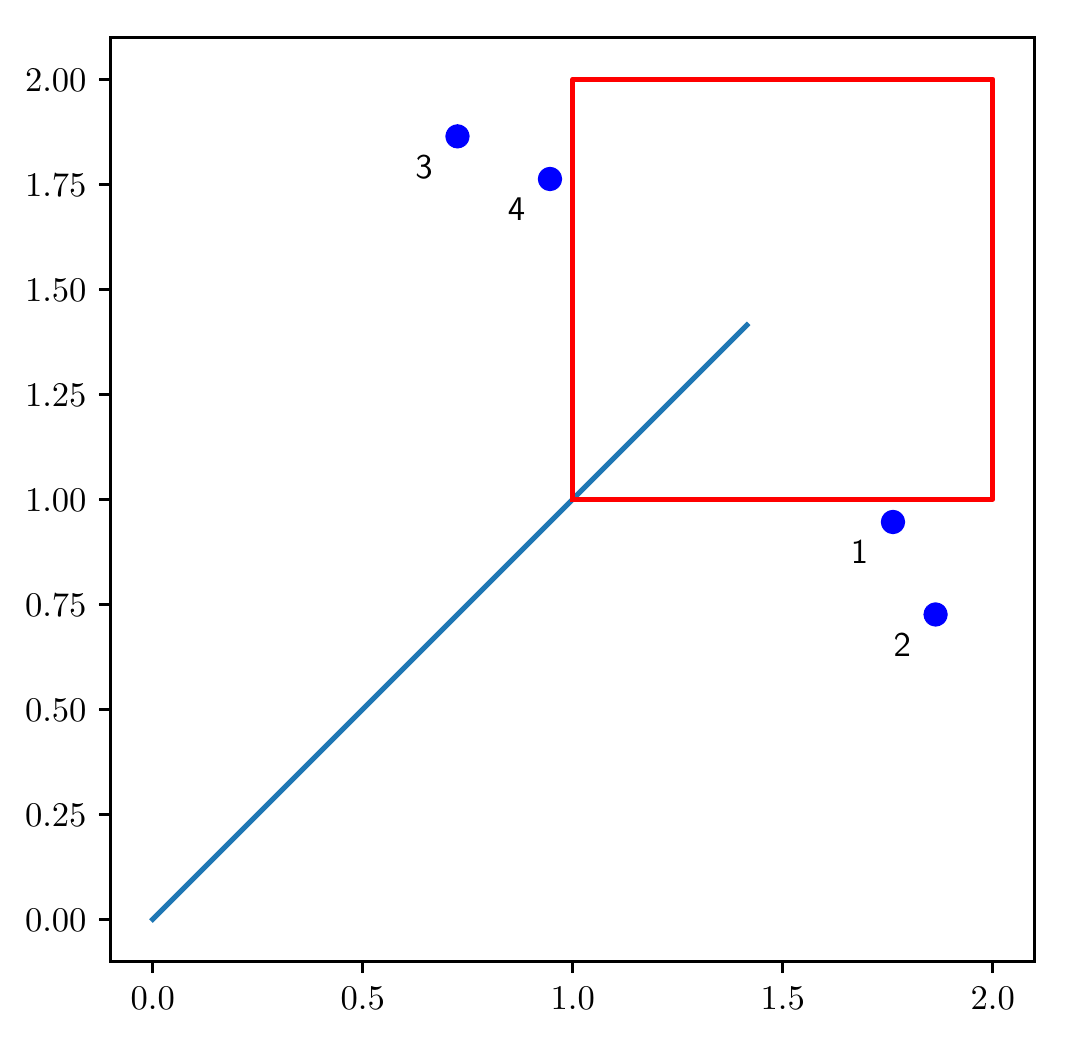}
  \caption{1R planar arm, and estimates of end-effector collisions.}
  \label{fig:box_1r}
\end{figure}

Consider a 1R robot arm with base at the origin, and a single link of length 2, shown in Figure~\ref{fig:box_1r}. The configuration $\bq$ is the angle $\theta$; let the initial configuration be $\theta = \pi/4$. Constrain the endpoint of the arm to lie in a square region with vertices $\bo = ((1, 1), (2, 1), (2, 2), (1, 2))$.
The end effector coordinates are
\begin{equation}
  \bp(\bq) = (2 \cos \theta, 2 \sin \theta).
\end{equation}

There are four distance functions:
\begin{eqnarray}
  d_1 & = & \bp_y - \bo_{1y} = 2\sin{\theta} - 1 \\
  d_2 & = & -(\bp_x - \bo_{2x}) = -2\cos{\theta} + 2 \\
  d_3 & = & -(\bp_y - \bo_{3y} = -2\sin{\theta} + 2 \\
  d_4 & = & \bp_x - \bo_{4x} = 2\cos{\theta} - 1,
\end{eqnarray}
corresponding to distances from the bottom, right, top, and left walls. Computing the partial derivatives with respect to $\theta$,
\begin{equation}
    J_d(\bq) \dot \bq + \bd_0 =
    \begin{pmatrix}
    2 \cos{\theta} \\
    2 \sin{\theta} \\
    -2 \cos{\theta} \\
    -2 \sin{\theta}
  \end{pmatrix}
  \dot \bq + \bd_0 \ge 0
\end{equation}

\begin{equation}
    \begin{pmatrix}
    \sqrt{2} \\
    \sqrt{2} \\
    -\sqrt{2} \\
    -\sqrt{2}
  \end{pmatrix}
  \dot \bq +
  \begin{pmatrix}
    \sqrt{2} - 1\\
    2 - \sqrt{2}\\
    2 - \sqrt{2}\\
    \sqrt{2} - 1
  \end{pmatrix}
  \ge 0.
\end{equation}
Candidate boundary values for values for $\dot \bq$, or equivalently, $\Delta \theta$, are $\approx (-0.29, -0.41, 0.41, 0.29)$.
The value $\Delta \theta = -0.29$ corresponds to collision with the bottom wall, and the value $\Delta \theta = 0.29$ corresponds to collision with the left wall; these would be the first collisions to occur. These values are of course approximate, due to the linearization of $J$ around the initial configuration.

\section{Flexibility analysis using linear programming}

We discover approximate extreme configurations of very large 2D systems of loosely connected rigid bodies by solving the linear program
\begin{equation}
  \label{eq:jacobian-linear-program}
  \begin{aligned}
  & \underset{\dot q}{\text{max}}
  & & \bc^T \dot \bq \\
      & \text{subject to} & &  J(q)\dot \bq + \bd_0 \ge 0,
  \end{aligned}
\end{equation}
where $\bc$ is a vector of weights. The choice of $\bc$ allows us to tune the direction we wish to displace elements.

\begin{figure}
  \begin{subfigure}[t]{0.22\textwidth}
      \centering
      \includegraphics[width=1.4in]{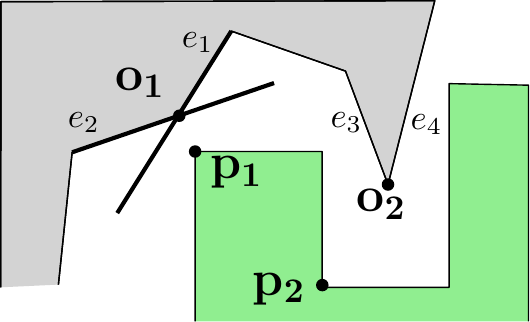}
      \caption{The edge-vertex distance constraints limit the valid motions in a small convex region. }
      \label{fig:distance-function-issue}
  \end{subfigure}
  ~
  \begin{subfigure}[t]{0.22\textwidth}
  \centering
  \includegraphics[width=1.6in]{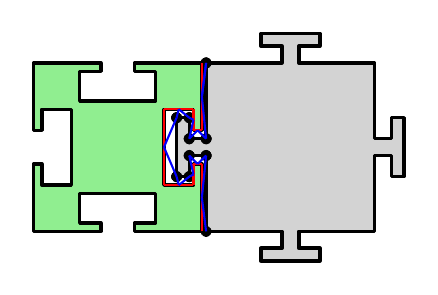}
    \caption{The distance constraints generated for a pair of rigid bodies. Blue line segments are vectors from green vertices to red edges.}
  \end{subfigure}
  \caption{Local distance functions model the free space.}
  \label{fig:distance-functions}
\end{figure}

We use signed distance functions between vertices on one body and edges on another to simply model the permissible local motions of the bodies. For each pair of bodies in the structure, we choose one body to provide the edges, and one body to provide vertices, as shown in Figure~\ref{fig:distance-function-issue}. Since the linearized analysis is only valid for local motions of the bodies, only edges and points that are initially near one another are potential sources of collision; we choose a small positive value $\epsilon$ and select edge/vertex pairs that are initially closer than this value.

Let the configurations of the current pair of bodies under consideration be $\bq_1 = (x_1, y_1, \theta_{1})$ and $\bq_2 = (x_2, y_2, \theta_2)$. For each object pair, we expect there to be many distance functions, representing the distances of vertices from edges over some region of near-contact between the bodies. For simplicity, consider a distance function $d_{ij}$ such that object 1 provides the edge, and object 2 provides the vertex. Let $\bn$ be the outwards-pointing normal from the edge and $\bo$ be the origin. Then
\begin{equation}
\label{equation:high-level-signed-distance}
\bd_{ij}(\bq_1, \bq_2) = \bn(\bq_1) \cdot (\bp(\bq_2) - \bo(\bq_1)).
\end{equation}

Let the length of the edge
be $\ell$, the first and second endpoints of the edge
be $e_0$ and $e_1$, the distance from the origin of object 1 to the endpoints of the edges be $r_{e_0}$ and $r_{e_1}$ and the angle from the x axis in the local frame of object 1 to the edge endpoints be $\alpha_{e_0}$ and $\alpha_{e_1}$. Let the distance from $\bp$ to the origin in the local frame of object 2 be $r_p$ and the angle from the x axis to $p$ be $\alpha_p$. To make the equations more readable, we define the helper variables
$s_{e_1} = \sin{(\theta_1 + \alpha_{e_1})}$,
$c_{e_1} = \cos{(\theta_1 + \alpha_{e_1})}$,
$s_{e_0} = \sin{(\theta_0 + \alpha_{e_0})}$,
$c_{e_0} = \cos{\theta_1 + \alpha_{e_0}}$,
$a = \frac{1}{\ell}$. Then
\begin{align}
    \bp(\bq_2) & = \begin{pmatrix}
      x_2 + r_p c_p \\
      y_2 + r_p s_p
    \end{pmatrix} \\
    \bn(\bq_1) & =
    a
    \begin{pmatrix}
        r_{e_1} s_{e_1} - r_{e_0} s_{e_0}  \\
        -r_{e_1} c_{e_1} + r_{e_0} c_{e_0}
    \end{pmatrix} \\
    \bo(\bq_1) & = \begin{pmatrix}
        x_1 + r_{e_0} c_{e_0} \\
        y_1 + r_{e_0} s_{e_0}
    \end{pmatrix}.
\end{align}

The non-zero entries of each row of the Jacobian are computed using the gradients of the distance function, substituting the appropriate blocks for blocks 1 and 2:
\resizebox{0.98\linewidth}{!}{
    \begin{minipage}{\linewidth}
        \begin{align}
            \partial_{x_1} \bd_{ij} & = a \left(- r_{e_0} s_{e_0} + r_{e_1} s_{e_1}\right) \nonumber \\
            \partial_{y_1} \bd_{ij}  & = a \left(c_{e_0} r_{e_0} - c_{e_1} r_{e_1}\right) \nonumber \\
            \partial_{\theta_1} \bd_{ij} & = a( c_{e_0} r_{e_0} \left(c_{e_0} r_{e_0} - c_{e_1} r_{e_1}\right) \nonumber \\
            & -  r_{e_0} s_{e_0} \left(- r_{e_0} s_{e_0} + r_{e_1} s_{e_1}\right) \nonumber \\
            & +  \left(- c_{e_0} r_{e_0} + c_{e_1} r_{e_1}\right) \left(- c_{p} r_{p} + c_{e_0} r_{e_0} + x_{1} - x_{2}\right) \nonumber \\
            & +  \left(- r_{e_0} s_{e_0} + r_{e_1} s_{e_1}\right) \left(- r_{p} s_{p} + r_{e_0} s_{e_0} + y_{1} - y_{2}\right)) \nonumber \\
            \partial_{x_2} \bd_{ij} & = - a \left(- r_{e_0} s_{e_0} + r_{e_1} s_{e_1}\right) \nonumber \\
            \partial_{y_2} \bd_{ij} & = - a \left(c_{e_0} r_{e_0} - c_{e_1} r_{e_1}\right) \nonumber \\
            \partial_{\theta_2} \bd_{ij} & = a (- c_{p} r_{p} \left(c_{e_0} r_{e_0} - c_{e_1} r_{e_1}\right) \nonumber \\
            & + r_{p} s_{p} \left(- r_{e_0} s_{e_0} + r_{e_1} s_{e_1}\right)) \nonumber \\
        \end{align}
    \end{minipage}
}

\subsection{Modeling convex corners}
\label{sec:convex_corners}

The distance-function approximation of the local configuration space is particularly bad for some object geometries. In Figure~\ref{fig:distance-function-issue}, point $\bp_1$ is closest to point $\bo_1$, a convex corner. Two distance functions are created, one for each of the extension into lines of the edges $e_1$ and $e_2$. Maintaining these constraints unnecessarily restricts $\bp_1$;  $\bp_1$ will remain in the polygonal region defined by the extensions of $e_1$ and $e_2$. This problem seems fundamental. Rows of the Jacobian express an {\em and} relationship; all constraints must be satisfied. But in the example, it is enough that $\bp_1$ be on the ``correct" side of only one of the extended edges.

If only one of the nearby vertices is convex, the problem is easily solvable. For example, in Figure~\ref{fig:distance-function-issue}, points $\bo_2$ and $\bp_2$ may be swapped, so that we compute the distance of a point relative to a concave corner.  To mitigate the problem in the case where both corners are convex, we may take a simple, though not entirely satisfactory, approach. Take the normals of each edge, and average them, yielding a half-plane constraint that at least allows $\bp_1$ to cross over the extended edges.
One promising avenue for a deeper exploration of this issue is formulation as a Linear Complementarity Problem (LCP), allowing {\em or} relationships between constraints.

\subsection{Time-stepping and re-enforcement of constraints}

Solutions to the linear program in Equation~\ref{eq:jacobian-linear-program} are extreme vertices of the constraint polyhedron. Because of the linearization around the initial configuration, the constraints may be violated when the resulting solution is used to compute a new configuration.

A common approach for dealing with truncation error in finite difference methods is to find the net change over several time steps. Although there are sophisticated ways to compute an optimal time step for finite difference methods, for this problem, the cost of computing the linear program solution far outweighs the cost of Euler-step integration and forward kinematics distance computation. We take a simple approach, and do a linear or binary search for a time step, multiplying the displacement vector $\Delta \bq$ by an increasingly larger scalar until the maximum distance constraint violation exceeds a user-defined threshold. After a time step is found and applied, a new linear program may be formulated and solved around the new configuration $\bq$. Usefully, the new linear program re-enforces the constraints, potentially taking a backward step with $\Delta \bq$. This means that error does not accumulate across time steps.

\section{Example problems}

In this section, we present some informal examples -- preliminary work that suggests interesting applications.

\subsection{Structure and block design problems}
The linear optimization approach may be fast enough for rapid consideration of different potential designs for a structure, including the number and locations of blocks, and on the the geometry of individual blocks, including the tightness of the joints.

\begin{figure}
  \centering
  \includegraphics[width=2.9in]{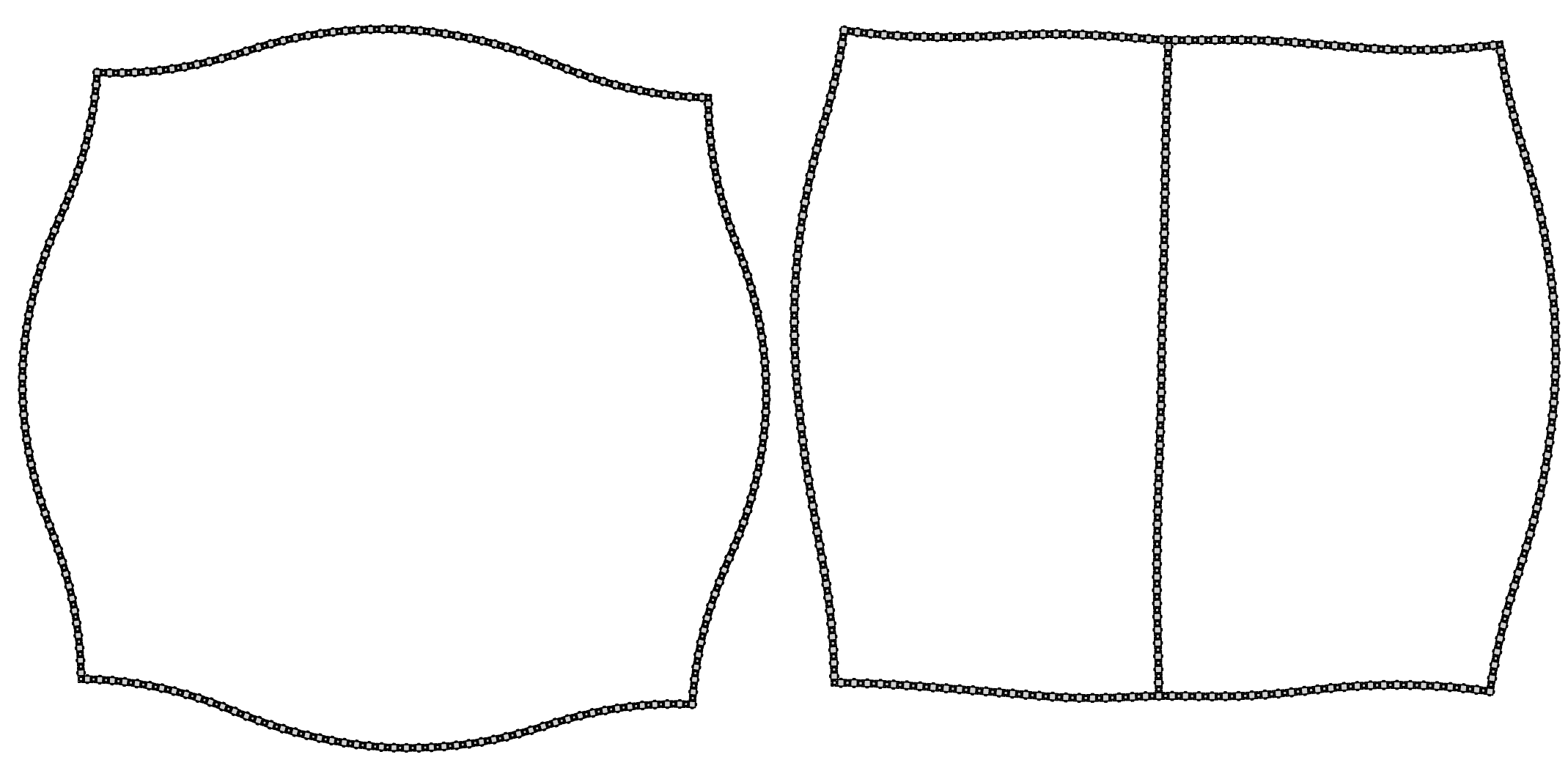}
    \caption{Example of adding a cross beam into a structure at a point of maximum flex.}
  \label{fig:topology-example}
\end{figure}

As an example, we consider how to add blocks to brace a structure and limit maximum flex. Figure~\ref{fig:topology-example} shows an example of adding such a beam to a structure; $\bc$ was chosen to maximize radial flex of each block outwards from the center. We find the pair of mutually visible vertices which has changed the most in the predicted configuration of maximum flex, an $O(n^2)$ operation for $n$ vertices. In this example, this approach suggested adding a vertical cross-beam of blocks, which we did by hand. In a completely automated algorithm, structural limitations of the blocks would need to be taken into account when selecting a cross beam.

\begin{figure}
  \centering
  \includegraphics[width=3.2in]{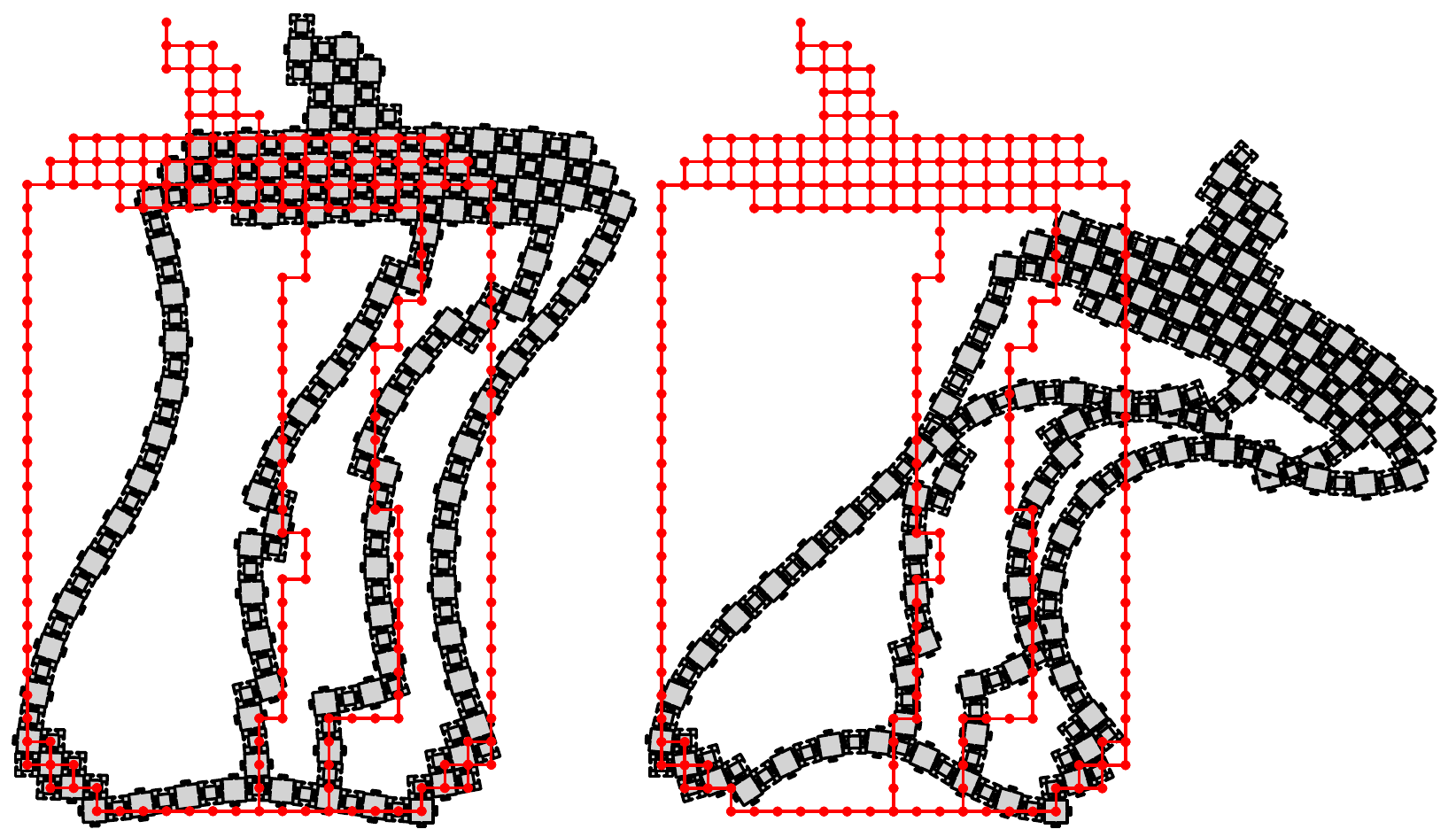}
    \caption{Crushing a soda can with tight and loose joints. Red polygons denote the initial configuration.}
  \label{fig:max-tolerance-example}
\end{figure}

The linear programming approach can also be used to explore joint geometry. Looser joints simplify assembly; if the joints in Figure~\ref{fig:chair} are too tight, the chair cannot be assembled due to limits on the precision of assembly and fabrication. However, if joints are too loose, the structure will flex unacceptably, particularly if there is wear on the connectors over time. Figure~\ref{fig:max-tolerance-example} (right) shows a planar example of the soda can with loosened joints, with flex computed using linear programming.

One simple strategy to explore joint tolerance is to parameterize the tightness of a joint with a single value and binary search for maximum tolerance. To simulate such a process, we utilize the Clipper library~\cite{Johnson2014-clipper-library} to simulate loosening joints by insetting the boundary of the rigid bodies. A more general approach might choose several parameters to describe joint geometry, and search over this parameter space.

\subsection{Flock formations}

\begin{figure}
  \centering
  \includegraphics[height=1.9in]{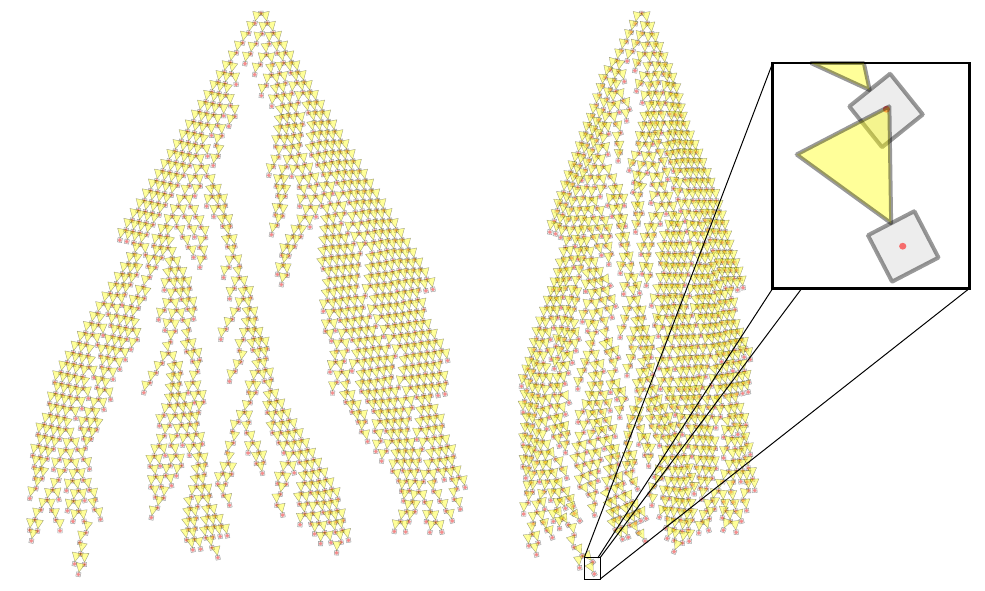}
    \caption{A linear programming solution for a flock of 1024 robots which must maintain sensor contact squeezing together to fit through a doorway or hallway.}
  \label{fig:flock-example}
\end{figure}

Figure~\ref{fig:flock-example} shows a flock of 1024 robots; the magnified inset shows the geometry. Gray square robots are forbidden from physical collision, and the yellow cone shows a requirement that each robot's camera must maintain view of a marker (red dot) on the robot in front of it.  We can drive the flock into interesting configurations by selecting an objective function. Figure~\ref{fig:flock-example} shows an example: driving the diffuse flock into a tighter configuration (perhaps so that the robots can pass through a doorway) by finding a displacement that moves all of the robots toward the x value of the leader robot.

We add field-of-view constraints for each robot except the leader, and collision constraints that require that vertices of each robot do not cross the half planes described by the edges of its five nearest robots.  We added a constraining square around the leader at the tip of the tree so that the constraint polyhedron is bounded. Large rotations are poorly approximated by the linear method, so we place an arbitrary limit on the rotation displacement of each robot in a time step, using auxiliary linear constraints. After each configuration update, we re-select distance constraints between swarm neighbors.

\subsection{Unbounded separation and (dis-)assembly planning}
\label{subsec:separation}

\begin{figure}
  	\centering
    \includegraphics[width=2.8in]{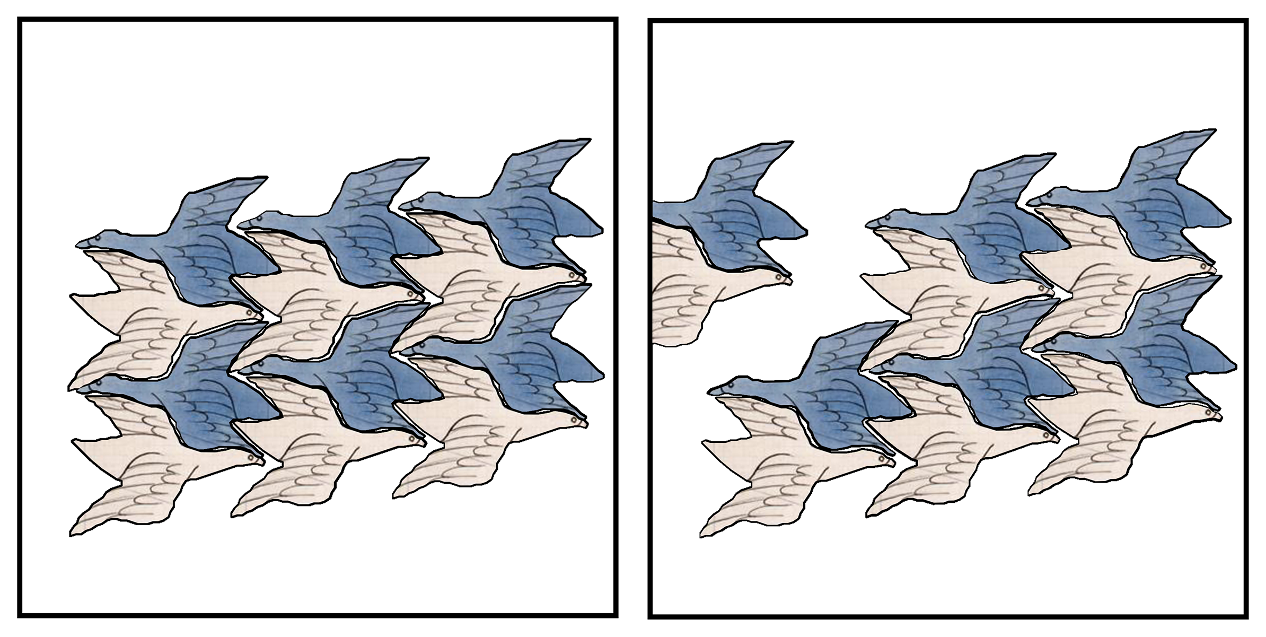}
    \caption{A small time step in a direction of separation.}
    \label{fig:separation-example}
\end{figure}

The classic assembly problem~\cite{HalperinLW00,SnoeyinkS94} is to discover motions that separate or assemble a collection of rigid bodies. For simple versions of this problem, we might like to discover a velocity direction $\dot \bq$ for which the linear constraints we formulated are unbounded. With some minor modifications, our approach is able to discover such a motion.

Linear program solvers are capable of detecting whether the feasible polyhedron is unbounded in the direction of a given cost vector; in contrast, we would like to discover such a cost vector automatically. Our approach is based on the observation that for almost every non-zero vector, the linear sum of the elements is either positive or negative, but not zero. We may compute the sum of the $x$ and $y$ elements of $\dot \bq$ by adding a row of the form $(1, 1, 0, 1, 1, 0, \dots)$ to $J$. We may constrain that sum to be very large, by adding an additional large element $k$ to $\bd_0$. Choose the objective function $\bc$ arbitrarily. We must also upper bound the motion so that the solution is not unbounded; we add a row $(-1, -1, 0, \ldots)$ and an element $-2k$ to $\bd_0$.

If a solution is found to this linear program, then the resulting $\dot \bq$ removes at least one block far enough from the assembly that it is unconstrained, allowing unbounded motion. If not, then we may look for negative motions by changing the signs on the last two elements of $\bd_0$. If both of these linear programs are infeasible, then the only separating motions must be such that the sum of the $x$ and $y$ velocity elements is exactly $0$. Our study of this approach is preliminary; Figure~\ref{fig:separation-example} shows an example of a direction of separation found.

\section{Evaluation and comparisons}

The size of the linear program depends on the number of blocks, the complexity of their shape, and the ways in which they are connected; the number of time-step iterations depends on the flexibility of the structure with respect to angular motions in configuration space. While there do not appear to be existing competing methods to solve exactly the problem under consideration, we explore the time costs for various problem sizes to serve as a baseline for future comparison.

For $n$ blocks with one block held fixed, the Jacobian has $3(n-1)$ columns and $c(n-1)$ rows, if $c$ is the average number of distance constraints generated for each block. However, the matrix is quite sparse, which may reduce memory and computational costs of solution; there are only six non-zero entries per row, yielding $O(n)$ non-zero entries in the matrix. We omit formal $O()$ asymptotic run-time analysis of the solution, since linear programming techniques are standard.

\begin{table}
  \begin{tabular}{c c c c}
    Rigid Bodies & Jacobian Size & Iterations & Runtime (seconds) \\ \hline
    36 & 666 x 105 & 5 & 0.358 \\
    50 & 907 x 147 & 8 & 0.363 \\
    90 & 2220 x 267 & 7 & 1.621 \\
    153 & 3147 x 456 & 8 & 1.866 \\
    223 & 5273 x 666 & 5 & 2.987 \\
    332 & 6309 x 993 & 5 & 2.761 \\
    392 & 10932 x 1173 & 8 & 10.649 \\
    396 & 7326 x 1185 & 4 & 2.002 \\
    688 & 14615 x 2061 & 4 & 5.951 \\
    1703 & 52655 x 5106 & 14 & 233.341 \\
    2497 & 66363 x 7488 & 7 & 166.582 \\
  \end{tabular}
  \caption{Performance results for several structures, using Gurobi with sparse matrices.}
  \label{table:performance-results}
\end{table}

In Table~\ref{table:performance-results} we show the result of tests on several systems of rigid bodies of varying size. For each structure, we report the amount of time and number of iterations required for our time stepping procedure to converge. The run time of our approach is dominated by the solution of the constraint
Jacobian linear program. In our experiments, we found that certain instances were especially hard for the linear program solver. For instance, the 392 rigid body structure takes five times as long the 396 body structure to solve and two times as long as the 688 body structure. The 392 body structure is a very dense structure, making the placement of each rigid body dependent on a larger number of other rigid bodies than in less dense structures.

\section{Limitations and future work}

We presented a simple linear-constraint method for computing the motion of a loosely-connected chain of rigid bodies. Like robot kinematics formulations, the approach is geometric, and does not model dynamics and contact. This is both a strength and a weakness; dynamics simulators may provide realistic motions, but the linear constraints describe a space of possible motion of the system, allowing fast and interesting optimizations. The linear constraint method may also be more useful for a worst-case analysis; just because a simulator provides a trajectory does not mean that trajectory will occur in the real world.

The linear-constraint method assumes that the configuration space is tight enough that linearization of the change in distance functions with respect to configuration-space motion is not too inaccurate.
For more flexible systems, the computed motions violate the distance constraints. Repeated enforcement of the constraints by time-stepping and re-solving the linear program gives results that seem empirically reasonable, but there is much to be done to put this approach on firmer mathematical footing, perhaps by analyzing Taylor series approximations~\cite{Duistermaat2010-taylor-polynomial-many-variables}.

The use of a linear objective function is also limiting.
For example, while we can analyze separability of objects (Section~\ref{subsec:separation}), there is little control over which separating motion is discovered.We might like to separate objects in an assembly one at time (if we have only one robot arm), or simultaneously, for speed; it is unclear how these preferences might be encoded with linear objective functions.

The use of the union of edge-vertex distance constraints to approximate the local configuration space also needs further study; as pointed out in Section~\ref{sec:convex_corners}, convex corners of objects pose a particular problem when used as edges for the distance function. Extension to 3D, an obvious next step for the work, seems mostly straight-forward, but we expect expressing the geometry of convex vertices, saddles, and ridges using a union of linear constraints to be more problematic than in the 2D case.

\renewcommand*{\bibfont}{\small}
\printbibliography

@STRING{IEEE_J_RA         = "{IEEE} Trans. Robot. Autom."}

@STRING{IEEE_J_RO         = "{IEEE} Trans. Robot."}

@STRING{IJRR = {Int. J. Robot. Res.}}

@STRING{ICRA = {Proc. {ICRA}}}

@STRING{IROS = {Proc. {IROS}}}

@STRING{WAFR = {Proc. WAFR}}

@inproceedings{Zhang2018-interlocking,
  author = {Yinan Zhang and Devin Balkcom},
  title     = {Interlocking block assembly},
  booktitle = wafr,
  year      = {2018},
  month = dec,
}

@inproceedings{Zhang2016a,
  author = {Yinan Zhang and Devin Balkcom},
  title = {Interlocking structure assembly with voxels},
  booktitle = iros,
  year = {2016}
}

@article{Balkcom2002b,
  author    = {Devin Balkcom and
               Jeffrey C. Trinkle},
  title     = {Computing wrench cones for planar rigid body contact tasks},
  journal   = ijrr,
  year      = {2002},
  volume    = {21},
  number    = {12},
  pages     = {1053--1066},
}

@misc{lensgraf2019puzzleflex,
    title={PuzzleFlex: kinematic motion of chains with loose joints},
    author={Samuel Lensgraf and Karim Itani and Yinan Zhang and Zezhou Sun and Yijia Wu and Alberto Quattrini Li and Bo Zhu and Emily Whiting and Weifu Wang and Devin Balkcom},
    year={2019},
    eprint={1906.08708},
    archivePrefix={arXiv},
    primaryClass={cs.RO}
}

@inproceedings{Li1993,
author = {Li, Zhenyu and Milenkovic, Victor},
title = {A Compaction Algorithm for Non-convex Polygons and Its Application},
booktitle = {Proceedings of the Ninth Annual Symposium on Computational Geometry},
series = {SCG '93},
year = {1993},
pages = {153--162}
}

@inproceedings{Hauser2018,
author = "Kris Hauser",
title = "Semi-Infinite Programming for Trajectory Optimization with Nonconvex Obstacles",
booktitle = "Workshop on the Algorithmic Foundations of Robotics (WAFR)",
month = 12,
year = 2018
}

@inproceedings{Choi02,
author = "Kwang-Jin Choi and Hyeong-Seok Ko",
title = "Stable but Responsive Cloth",
booktitle = "SIGGRAPH",
year = 2002,
}

@book       { Rubin2000,
author  =   "M.B. Rubin",
title   =   "Cosserat Theories:  Shells, Rods, and Points",
publisher=  "Kluwer Academic Publishers",
year    =   2000,
}

@inproceedings{Baraff00,
 author = {Baraff, David and Witkin, Andrew},
 title = {Large Steps in Cloth Simulation},
 booktitle = {Proceedings of the 25th Annual Conference on Computer Graphics and Interactive Techniques},
 series = {SIGGRAPH '98},
 year = {1998},
 isbn = {0-89791-999-8},
 pages = {43--54},
 numpages = {12},
 url = {http://doi.acm.org/10.1145/280814.280821},
 doi = {10.1145/280814.280821},
 acmid = {280821},
 publisher = {ACM},
 address = {New York, NY, USA},
}

@article{James2003,
    author    = {Doug L. James and Dinesh K. Pai},
    title     = {Multiresolution {G}reen's function methods for interactive simulation of large-scale elastostatic objects},
    journal   = {ACM Trans. Graph.},
    volume    = 22,
    number    = 1,
    year      = 2003,
    issn      = {0730-0301},
    pages     = {47--82},
    doi       = {http://doi.acm.org/10.1145/588272.588278},
    publisher = {ACM Press},
    address   = {New York, NY, USA},
}

@inproceedings{Pai2002,
author = "Dinesh K. Pai",
title = "{STRANDS}: Interactive Simulation of Thin Solids using Cosserat Models",
booktitle = "Eurographics",
year = 2002,
}

@article{HalperinLW00,
  author    = {Dan Halperin and
               Jean-Claude Latombe and
               Randall H. Wilson},
  title     = {A General Framework for Assembly Planning: The Motion Space Approach},
  journal   = {Algorithmica},
  volume    = {26},
  number    = {3-4},
  pages     = {577--601},
  year      = {2000},
}

@article{SnoeyinkS94,
  author    = {Jack Snoeyink and
               Jorge Stolfi},
  title     = {Objects that Cannot Be Taken Apart with Two Hands},
  journal   = {Discrete and Computational Geometry},
  volume    = {12},
  pages     = {367--384},
  year      = {1994},
}

@inproceedings{ShahrokhiMB17,
  author    = {Shiva Shahrokhi and
               Arun Mahadev and
               Aaron T. Becker},
  title     = {Algorithms for shaping a particle swarm with a shared input by exploiting non-slip wall contacts},
  booktitle = iros,
  pages     = {4304--4311},
  year      = {2017}
}

@ARTICLE{TTPrs01,
  AUTHOR = {J.C. Trinkle and J.A. Tzitzouris and J.S. Pang},
  TITLE = {Dynamic Multi-Rigid-Body Systems with Concurrent Distributed Contacts: Theory and Examples},
  PUBLISHER = {The Royal Society},
  JOURNAL = {Philosophical Transactions: Mathematical, Physical, and Engineering Sciences},
  MONTH = {December},
  SERIES = {A},
  VOLUME = 359,
  NUMBER = 1789,
  PAGES = {2575--2593},
  YEAR = 2001,
}

@INCOLLECTION{STsiam97,
  AUTHOR = {D.E. Stewart and J.C. Trinkle},
  TITLE = {Dynamics, Friction, and Complementarity Problems},
  BOOKTITLE = {Complementarity and Variational Problems},
  EDITOR = {M.C. Ferris and J.S. Pang},
  PUBLISHER = {SIAM},
  YEAR = 1997,
  PAGES = {425--439},
}

@article{SacksBM17,
  author    = {Elisha Sacks and
               Nabeel Butt and
               Victor Milenkovic},
  title     = {Robust free space construction for a polyhedron with planar motion},
  journal   = {Computer-Aided Design},
  volume    = {90},
  pages     = {18--26},
  year      = {2017},
}

@article{RodriguezMF12,
  author    = {Alberto Rodriguez and
               Matthew T. Mason and
               Steve Ferry},
  title     = {From caging to grasping},
  journal   = ijrr,
  volume    = {31},
  number    = {7},
  pages     = {886--900},
  year      = {2012},
}

@inproceedings{makita2008,
  title={3D multifingered caging: Basic formulation and planning},
  author={Makita, Satoshi and Maeda, Yusuke},
  booktitle=IROS,
  pages={2697--2702},
  year={2008}
}

@article{vahedi2008caging,
  title={Caging polygons with two and three fingers},
  author={Vahedi, Mostafa and van der Stappen, A Frank},
  journal=IJRR,
  volume={27},
  number={11-12},
  pages={1308--1324},
  year={2008}
}

@inproceedings{erickson2003capturing,
  title={Capturing a convex object with three discs},
  author={Erickson, Jeff and Thite, Shripad and Rothganger, Fred and Ponce, Jean},
  booktitle=ICRA,
  volume={2},
  pages={2242--2247},
  year={2003},
}

@inproceedings{rimon1996caging,
  title={Caging 2D bodies by 1-parameter two-fingered gripping systems},
  author={Rimon, Elon and Blake, Andrew},
  booktitle=ICRA,
  pages={1458--1464},
  year={1996}
}

@article{allen2015two,
  title={Two-finger caging of polygonal objects using contact space search},
  author={Allen, Thomas F and Burdick, Joel W and Rimon, Elon},
  journal=IEEE_J_RO,
  volume={31},
  number={5},
  pages={1164--1179},
  year={2015},
}

@book{Reuleaux1876,
  author = {Franz Reuleaux},
  title = {The kinematics of machinery},
  year = {1876}
  }

@article{Connelly2003-carpenters,
  AUTHOR        = {Robert Connelly and Erik D. Demaine and G\"unter Rote},
  TITLE         = {Straightening Polygonal Arcs and Convexifying Polygonal
                   Cycles},
  JOURNAL       = {Discrete and Computational Geometry},
  JOURNALURL    = {http://link.springer.de/link/service/journals/00454/},
  VOLUME        = 30,
  NUMBER        = 2,
  MONTH         = sep,
  YEAR          = 2003,
  PAGES         = {205--239},
  }

@inproceedings{Berenson2013-deformable,
  author    = {Dmitry Berenson},
  title     = {Manipulation of deformable objects without modeling and simulating
               deformation},
  booktitle = IROS,
  pages     = {4525--4532},
  year      = {2013},
}

@article{Bicchi2000-manipulability,
  author    = {Antonio Bicchi and
               Domenico Prattichizzo},
  title     = {Manipulability of cooperating robots with unactuated joints and closed-chain
               mechanisms},
  journal   = IEEE_J_RA,
  volume    = {16},
  number    = {4},
  pages     = {336--345},
  year      = {2006},
  month     = aug
}

@inproceedings{Kim1998-manipulability,
  author    = {Sungbok Kim},
  title     = {Adjustable Manipulability of Closed-Chain Mechanisms through Joint
               Freezing and Joint Unactuation},
  booktitle = ICRA,
  pages     = {2627--2632},
  year      = {1998},
}

@inproceedings{Park1998-manipulability,
  author    = {Frank C. Park and
               Jin Wook Kim},
  title     = {Manipulability and Singularity Analysis of Multiple Robot Systems:
               a Geometric Approach},
  booktitle = ICRA,
  pages     = {1032--1037},
  year      = {1998},
}

@article{Chiacchio1991-manipulability,
  author    = {Pasquale Chiacchio and
               Stefano Chiaverini and
               Lorenzo Sciavicco and
               Bruno Siciliano},
  title     = {Global task space manipulability ellipsoids for multiple-arm systems},
  journal   = IEEE_J_RA,
  volume    = {7},
  number    = {5},
  pages     = {678--685},
  year      = {1991},
}

@inproceedings{Werfel2006-mobile-robot-construction,
    author    = {Justin Werfel and
                 Yaneer Bar-Yam and
                 Daniela Rus and
                 Radhika Nagpal},
    title     = {Distributed Construction by Mobile Robots with Enhanced Building Blocks},
    booktitle = ICRA,
    pages     = {2787--2794},
    year      = {2006}
}

@inproceedings{Eckenstein2017-acceptance-area-connectors,
    author    = {Nick Eckenstein and Mark Yim},
    title     = {Modular robot connector area of acceptance from configuration space obstacles},
    booktitle = IROS,
    pages     = {3550--3555},
    year      = {2017},
}

@article{Mishra1987-grasp-existence,
    author    = {Bhubaneswar Mishra and
                             Jacob T. Schwartz and
                                                Micha Sharir},
    title     = {On the Existence and Synthesis of Multifinger Positive Grips},
    journal   = {Algorithmica},
    volume    = {2},
    pages     = {541--558},
    year      = {1987},
}

@article{Makita2017-caging-survey,
  author    = {Satoshi Makita and
               Weiwei Wan},
  title     = {A survey of robotic caging and its applications},
  journal   = {Advanced Robotics},
  volume    = {31},
  number    = {19-20},
  pages     = {1071--1085},
  year      = {2017},
}

@article{Chase1991-survey-of-tolerance-analysis,
    author  = "Chase, Kenneth W.
    and Parki nson, Alan R.",
    title   = "A survey of research in the application of tolerance analysis to the design of mechanical assemblies",
    journal = "Research in Engineering Design",
    year    = "1991",
    month   = mar,
    volume  = "3",
    number  = "1",
    pages   = "23--37",
}

@article{Chase1996-geometric-tolerance-analysis,
    author    = {Kenneth W. Chase and Jinsong Gao and Spencer P. Magleby and Carl D. Sorensen},
    title     = {Including Geometric Feature Variations in Tolerance Analysis of Mechanical Assemblies},
    journal   = {IIE Transactions},
    volume    = {28},
    number    = {10},
    pages     = {795-807},
    year      = {1996},
}

@article{Revels2016-automatic-differentiation-in-julia,
    title   = {Forward-Mode Automatic Differentiation in Julia},
    author  = {{Revels}, J. and {Lubin}, M. and {Papamarkou}, T.},
    journal = {arXiv:1607.07892 [cs.MS]},
    year    = {2016},
}

@Inbook{Duistermaat2010-taylor-polynomial-many-variables,
    author    = "Duistermaat, J. J. and Kolk, J. A. C.",
    title     ="Taylor Expansion in Several Variables",
    bookTitle = "Distributions: Theory and Applications",
    year      = "2010",
    publisher = "Birkh{\"a}user",
    pages     = "59--63",
}

@misc{Johnson2014-clipper-library,
    author = "Angus Johnson",
    title  = "Clipper - an open source freeware library for
clipping and offsetting lines and polygons",
    yeah   = 2014,
    url    = "http://angusj.com/delphi/clipper.php"
}

@misc{gurobi,
  author = "Gurobi Optimization, LLC",
  title = "Gurobi Optimizer Reference Manual",
  year = 2018,
  url = "http://www.gurobi.com"
}

@MISC{powell-direct-search-optimization,
    author = {M. J. D. Powell},
    title = {Direct Search Algorithms for Optimization Calculations},
    year = {1998}
}

@Article{Zuo-jacobian-torsor-tolerance,
author="Zuo, XiaoYan
and Li, Beizhi
and Yang, Jianguo
and Jiang, Xiaohui",
title="Application of the Jacobian--torsor theory into error propagation analysis for machining processes",
journal="The International Journal of Advanced Manufacturing Technology",
year="2013",
volume="69",
number="5",
pages="1557--1568",
}

@InProceedings{Laperriere-jacobian-tolerance,
author="Laperri{\`e}re, Luc
and Lafond, Philippe",
editor="Batoz, Jean-Louis
and Chedmail, Patrick
and Cognet, Gerard
and Fortin, Cl{\'e}ment",
title="Modeling Dispersions Affecting Pre-Defined Functional Requirements of Mechanical Assemblies Using Jacobian Transforms",
booktitle="Integrated Design and Manufacturing in Mechanical Engineering '98",
year="1999",
}

@article{Nelder-Mead-simplex-optimization,
    author = {Nelder, J. A. and Mead, R.},
    title = "{A Simplex Method for Function Minimization}",
    journal = {The Computer Journal},
    volume = {7},
    number = {4},
    pages = {308-313},
    year = {1965},
    month = {01},
}

@article{hsieh2008decentralized,
  title={Decentralized controllers for shape generation with robotic swarms},
  author={Hsieh, M Ani and Kumar, Vijay and Chaimowicz, Luiz},
  journal={Robotica},
  volume={26},
  number={5},
  pages={691--701},
  year={2008},
  publisher={Cambridge University Press}
}

@inproceedings{o2014self,
  title={Self-assembly of a swarm of autonomous boats into floating structures},
  author={O'Hara, Ian and Paulos, James and Davey, Jay and Eckenstein, Nick and Doshi, Neel and Tosun, Tarik and Greco, Jonathan and Seo, Jungwon and Turpin, Matt and Kumar, Vijay and others},
  booktitle=icra,
  pages={1234--1240},
  year={2014}
}

@article{alonso2017multi,
  title={Multi-robot formation control and object transport in dynamic environments via constrained optimization},
  author={Alonso-Mora, Javier and Baker, Stuart and Rus, Daniela},
  journal=ijrr,
  volume={36},
  number={9},
  pages={1000--1021},
  year={2017},
  publisher={SAGE Publications Sage UK: London, England}
}

@article{rubenstein2014programmable,
  title={Programmable self-assembly in a thousand-robot swarm},
  author={Rubenstein, Michael and Cornejo, Alejandro and Nagpal, Radhika},
  journal={Science},
  volume={345},
  number={6198},
  pages={795--799},
  year={2014},
  publisher={American Association for the Advancement of Science}
}

@inproceedings{esposito2006maintaining,
  title={Maintaining wireless connectivity constraints for swarms in the presence of obstacles},
  author={Esposito, Joel M and Dunbar, Thomas W},
  booktitle=icra,
  pages={946--951},
  year={2006}
}

\end{document}